\pdfoutput=1

\documentclass[11pt]{article}

\usepackage[]{EMNLP2022/emnlp2022}

\usepackage{times}
\usepackage{latexsym}

\usepackage{graphicx}
\usepackage{amsmath}
\usepackage{amssymb}
\usepackage{booktabs}
\usepackage{mathtools}
\usepackage{kotex}
\usepackage{xcolor}         
\usepackage{xspace}
\usepackage{enumitem}
\usepackage{cancel}
\usepackage{multirow}
\usepackage{tabularx}
\usepackage{algorithm}
\usepackage{algpseudocode}

\usepackage[T1]{fontenc}

\usepackage[utf8]{inputenc}

\usepackage{microtype}

\usepackage{inconsolata}

\newcommand{\MI}{\text{MI}}
\newcommand{\XMI}{\text{XMI}}

\title{Specializing Multi-domain NMT via Penalizing Low Mutual Information}

\author{Jiyoung Lee$^{\dagger}$\thanks{\hspace{3pt} Work done during an internship at NAVER Corp.} , Hantae Kim$^{\ddagger}$, Hyunchang Cho$^{\ddagger}$ \\
    \textbf{Edward Choi$^{\dagger}$, and Cheonbok Park$^{\ddagger}$} \\
  $^{\dagger}$KAIST , $^{\ddagger}$Papago, NAVER Corp.\\
  \texttt{\{jiyounglee0523, edwardchoi\}@kaist.ac.kr} \\
  \texttt{\{hantae.kim,hyunchang.cho,cbok.park\}@navercorp.com} }
\begin{document}
\maketitle
\begin{abstract}
Multi-domain Neural Machine Translation (NMT) trains a single model with multiple domains. 
It is appealing because of its efficacy in handling multiple domains within one model.
An ideal multi-domain NMT should learn distinctive domain characteristics simultaneously, however, grasping the domain peculiarity is a non-trivial task. 
In this paper, we investigate domain-specific information through the lens of mutual information (MI) and propose a new objective that penalizes low MI to become higher.
Our method achieved the state-of-the-art performance among the current competitive multi-domain NMT models. 
Also, we empirically show our objective promotes low MI to be higher resulting in domain-specialized multi-domain NMT. 
\end{abstract}

\section{Introduction}
Multi-domain Neural Machine Translation (NMT) \citep{sajjad2017neural, farajian2017multi} has been an attractive topic due to its efficacy in handling multiple domains with a single model.
Ideally, a multi-domain NMT should capture both general knowledge (\textit{e.g.}, sentence structure, common words) and domain-specific knowledge (\textit{e.g.}, domain terminology) unique in each domain.
While the shared knowledge can be easily acquired via sharing parameters across domains \citep{kobus2016domain}, obtaining domain specialized knowledge is a challenging task. 
\citet{haddow2012analysing} demonstrate that a  model trained on multiple domains sometimes  underperforms the one trained on a single domain.
\citet{pham2021revisiting} shows that separate domain-specific adaptation modules are not sufficient to fully-gain specialized knowledge. 

\begin{figure}
    \centering
    \includegraphics[width=\columnwidth]{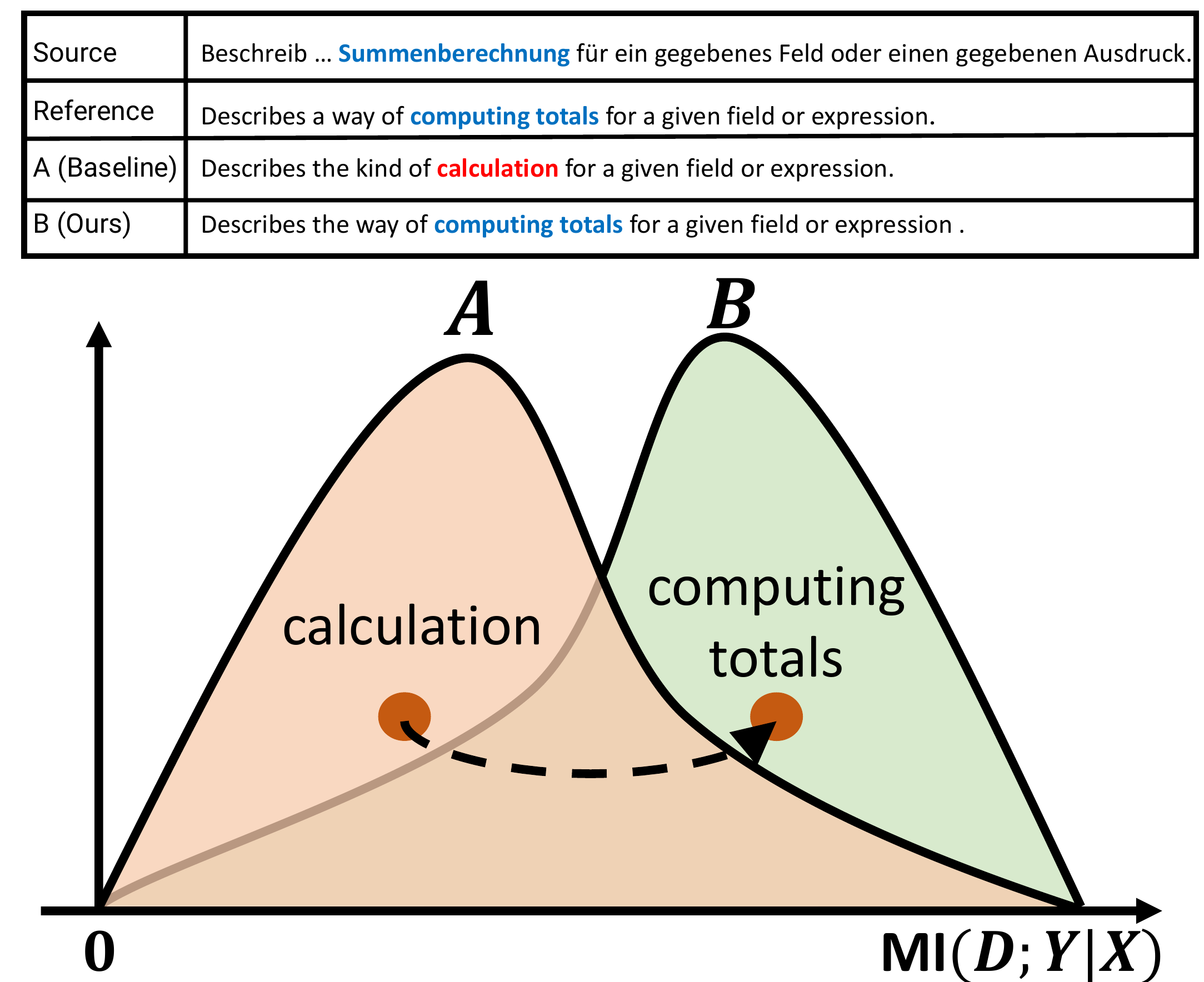}
    \caption{Overview of two models with different MI distributions. The example sentence is from IT domain. Model A mostly has low MI and Model B has large MI. For an identical sample, model A outputs a generic term `calculation' while model B properly maintains `computing totals'.}
    \label{fig:overview}
\end{figure}

In this paper, we reinterpret domain specialized knowledge from mutual information (MI) perspective and propose a method to strengthen it.
Given a source sentence $X$, target sentence $Y$, and corresponding domain $D$, the MI between $D$ and the translation $Y|X$ (\textit{i.e.}, $\MI(D;Y|X)$) measures the dependency between the domain and the translated sentence. 
Here, we assume that the larger $\MI(D;Y|X)$, the more the translation incorporates domain knowledge. 
Low MI is undesirable because it indicates the model is not sufficiently utilizing domain characteristics in translation.
In other words, low MI can be interpreted as a domain-specific information the model has yet to learn.
For example, as shown in Fig.~\ref{fig:overview}, we found that a model with low MI translates an IT term `computing totals' to the vague and plain term `calculation'.
However, once we force the model to have high MI, `computing totals' is correctly retained in its translation.
Thus, maximizing MI promotes multi-domain NMT to be domain-specialized.

Motivated by this idea, we introduce a new method that specializes multi-domain NMT by penalizing low MI.
We first theoretically derive $\MI(D;Y|X)$, and formulate a new objective that weights more penalty on subword-tokens with low MI.
Our results show that the proposed method improves the translation quality in all domains. 
Also, the MI visualization ensures that our method is effective in maximizing MI.
We also observed that our model performs particularly better on samples with strong domain characteristics.

The main contributions of our paper are as follows:
\begin{itemize}
  \item We investigate MI in multi-domain NMT and present a new objective that penalizes low MI to have higher value. 
  \item Extensive experiment results prove that our method truly yields high MI, resulting in domain-specialized model.
\end{itemize}

\section{Related Works}
\paragraph{Multi-Domain Neural Machine Translation}
Multi-Domain NMT focuses on developing a proper usage of domain information to improve translation. 
Early studies had two main approaches: injecting source domain information and adding a domain classifier.
For adding source domain information, \citet{kobus2016domain} inserts a source domain label as an additional tag with input or as a complementary feature.
For the second approach, \citet{britz-etal-2017-effective} trains the sentence embedding to be domain-specific  by updating using the gradient from the domain-classifier.

While previous work leverages domain information by injection or implementing an auxiliary classifier, we view domain information from MI perspective and propose a loss that promotes model to explore domain specific knowledge.

\paragraph{Information-Theoretic Approaches in NMT}
Mutual information in NMT is primarily used either as metrics or a loss function.
For metrics, \citet{bugliarello-etal-2020-easier} proposes cross-mutual information (XMI) to quantify the difficulty of translating between languages.
\citet{fernandes-etal-2021-measuring} modifies XMI to measure the usage of the given context during translation.
For the loss function, \citet{xu-etal-2021-bilingual} proposes bilingual mutual information (BMI) which calculates the word mapping diversity, further applied in NMT training.
\citet{zhang-etal-2022-conditional} improves the model translation by maximizing the MI between a target token and its source sentence based on its context. 

Above work only considers general machine translation scenarios.
Our work differs in that we integrate mutual information in multi-domain NMT to learn domain-specific information.
Unlike other methods that require training of an additional model, our method can calculate MI within a single model which is more computation-efficient.
\section{Proposed Method}
In this section, we first derive MI in multi-domain NMT. 
Then, we introduce a new method that penalizes low MI to have high value resulting in a domain-specialized model.

\subsection{Mutual Information in Multi-Domain NMT}
Mutual Information (MI) measures a mutual dependency between two random variables.
In multi-domain NMT, the MI between the domain ($D$) and translation ($Y|X$), expressed as $\MI(D;Y|X)$, represents how much domain-specific information is contained in the translation.
$\MI(D;Y|X)$ can be written as follows:
\begin{align}
    \MI(D;Y|X) &= \mathbb{E}_{D,X,Y} \bigg[ \log \frac{p(Y|X,D)}{p(Y|X)} \bigg].
\end{align}
The full derivation can be found in Appendix~\ref{sec:DAMI derivation}.
Note that the final form of $\MI(D;Y|X)$ is a log quotient of the translation considering domain and translation without domain.

Since the true distributions are unknown, we approximate them with a parameterized model~\citep{bugliarello-etal-2020-easier, fernandes-etal-2021-measuring}, namely the cross-MI (XMI).
Naturally, a generic domain-agnostic model (further referred to as \textit{general} and abbreviated as \textit{G}) output would be the appropriate approximation of $p(Y|X)$. A domain-adapted (further shortened as \textit{DA}) model output would be suitable for $p(Y|X,D)$. 
Hence, $\XMI(D;Y|X)$ can be expressed as Eq.~(\ref{approximated MI}) with each model output.
\begin{equation}
\begin{multlined}
    \XMI(D;Y|X) = \mathbb{E}_{D,X,Y} \bigg[ \log \frac{p_{\textit{DA}}(Y|X,D)}{p_{{G}}(Y|X)} \bigg]
    \label{approximated MI}
\end{multlined}
\end{equation}
    
\subsection{MI-based Token Weighted Loss}
To calculate XMI, we need outputs from both general and domain-adapted models. 
Motivated by the success of adapters \citep{houlsby2019parameter} in multi-domain NMT \citep{pham2021revisiting}, we assign adapters $\phi_1, \cdots \phi_N$ for each domain ($N$ is the total number of domains) and have an extra adapter $\phi_G$ for  general.
We will denote the shared parameter (\textit{e.g.}, self-attention and feed-forward layer) as $\theta$.
For a source sentence $x$ from domain $d$, $x$ passes the model twice, once through the corresponding domain adapter, $\phi_d$, and the other through the general adapter, $\phi_G$. 
Then, we treat the output probability from domain adapter as $p_{{DA}}$ and from general adapter as $p_{{G}}$.
For the $i^{th}$ target token, $y_i$ ,we calculate XMI as in Eq.~(\ref{delta_i}),
\begin{equation}
    {p(y_i | y_{<i}, x, \theta, \phi_d)} - {p(y_i | y_{<i}, x, \theta, \phi_G)}
    \label{delta_i}
\end{equation}
, where $y_{<i}$ is the target subword-tokens up to, but excluding $y_i$.
For simplicity, we will denote Eq.~(\ref{delta_i}) as $\XMI(i)$.
Low $\XMI(i)$ means that our domain adapted model is not thoroughly utilizing domain information during translation.
Therefore, we weight more on the tokens with low $\XMI(i)$, resulting in minimizing Eq.~(\ref{delta_i_loss}),
\begin{equation}
   \mathcal{L_\MI} = \sum_{i=0}^{n_T} (1-\XMI(i)) \cdot (1 - p(y_i | y_{<i}, x, \theta, \phi_d))
   \label{delta_i_loss}
\end{equation}
, where $n_T$ is the number of subword-tokens in the target sentence.

The final loss of our method is in Eq.~(\ref{final_objective}), where $\lambda_1$ and $\lambda_2$ are hyperparameters.
\begin{align}
    \mathcal{L_\text{DA}} &= -\sum_{i=0}^{n_T}\log(p(y_i | y_{<i}, x, \theta, \phi_d))\\
    \mathcal{L_\text{G}} &= -\sum_{i=0}^{n_T}\log(p(y_i | y_{<i}, x, \theta, \phi_G)) \\
    \mathcal{L} &= \mathcal{L_\text{DA}} + \lambda_1\mathcal{L_\text{G}} + \lambda_{2}\mathcal{L_\text{MI}}
    \label{final_objective}
\end{align}


\section{Experiments}
\subsection{Experiment Setting}
\paragraph{Dataset.}
We leverage the preprocessed dataset released by \citet{aharoni-goldberg-2020-unsupervised} consisting of five domains (IT, Koran, Law, Medical, Subtitles) available in OPUS \cite{tiedemann2012parallel, aulamo-tiedemann-2019-opus}. 
More details on the dataset and preprocessing are described in Appendix~\ref{sec:Experiment Details}.

\paragraph{Baseline.} We compare our method with the following baseline models:
(1) \textbf{Mixed} trains a model on all domains with uniform distribution,
(2) \textbf{Domain-Tag} \citep{kobus2016domain} inserts domain information as an additional token in the input,
(3) \textbf{Multitask Learning (MTL)} \citep{britz-etal-2017-effective} trains a domain classifier simultaneously and encourage the sentence embedding to encompass its domain characteristics,
(4) \textbf{Adversarial Learning (AdvL)} \citep{britz-etal-2017-effective} makes the the sentence embedding to be domain-agnostic by flipping the gradient from the domain classifier before the back-propagation,
(5) \textbf{Word-Level Domain Context Discrimination
(WDC)} \citep{zeng2018multi} integrates two sentence embedding which are trained by MTL and AdvL respectively,
(6) \textbf{Word-Adaptive Domain Mixing}\footnote{We conducted experiments using publicly available code.} \citep{jiang-etal-2020-multi-domain}, has domain-specific attention heads and the final representation is the combination of each head output based on the predicted domain proportion,
and (7) \textbf{Domain-Adapter} \citep{pham2021revisiting} has separate domain adapters \citep{houlsby2019parameter} and a source sentence passes through its domain adapters. This can be regarded as our model without general adapter and trained with $\mathcal{L}_{\text{DA}}$.

\subsection{Main Results}
\label{sec:main_experiment}
\begin{table*}[]
\begin{center}
\resizebox{0.8\linewidth}{!}{
\begin{tabular}{c|ccccc|cc}
\toprule
                                                                                        &  IT                                                                  & Koran                                                               & Law                                                                 & Medical                                                             & Subtitles                                                           & Average                                                           \\ \hline
\multirow{2}{*}{Mixed}                                                                                                          & 43.87{\scriptsize $\pm$0.505}                                                        & 20.31{\scriptsize $\pm$0.371}                                                        & 58.33{\scriptsize $\pm$0.474}                                                        & 55.19{\scriptsize $\pm$0.737}                                                        & 30.36{\scriptsize $\pm$0.424}                                                       & 41.61                                                             \\
                                                                                                                                                   & 62.00{\scriptsize $\pm$0.403}                                                       & 41.75{\scriptsize $\pm$0.343}                                                       & 73.41{\scriptsize $\pm$0.303}                                                       & 69.14{\scriptsize $\pm$0.346}                                                       & 45.73{\scriptsize $\pm$0.424}                                                       & 58.40                                                             \\ \hline
\multirow{2}{*}{Domain-Tag}                                                                                                     & 44.29{\scriptsize $\pm$0.142}                                                        & 20.44{\scriptsize $\pm$0.236}                                                        & 58.47{\scriptsize $\pm$0.275}                                                        & 55.39{\scriptsize $\pm$0.288}                                                        & 30.61{\scriptsize $\pm$0.220}                                                        & 41.84                                                             \\
                                                                                                                                             & 62.30{\scriptsize $\pm$0.111}                                                       & 41.75{\scriptsize $\pm$0.203}                                                        & 73.56{\scriptsize $\pm$0.190}                                                        & 69.28{\scriptsize $\pm$0.160}                                                        & 45.99{\scriptsize $\pm$0.268}                                                        & 58.58                                                             \\ \hline
\multirow{2}{*}{MTL}                                                                                         & 44.00{\scriptsize $\pm$0.298}                                                        & 20.40{\scriptsize $\pm$0.198}                                                        & 58.27{\scriptsize $\pm$0.327}                                                        & 55.24{\scriptsize $\pm$0.564}                                                        & 30.52{\scriptsize $\pm$0.478}                                                        & 41.69                                                             \\
                                                                                                                                           & 62.11{\scriptsize $\pm$0.169}                                                       & 41.78{\scriptsize $\pm$0.174}                                                        & 73.42{\scriptsize $\pm$0.197}                                                        & 69.16{\scriptsize $\pm$0.235}                                                        & 45.87{\scriptsize $\pm$0.316}                                                        & 58.47                                                             \\ \hline
\multirow{2}{*}{AdvL}                                                                                                    & 43.86{\scriptsize $\pm$0.167}                                                        & 20.33{\scriptsize $\pm$0.275}                                                        & 58.40{\scriptsize $\pm$0.195}                                                        & 55.56{\scriptsize $\pm$0.245}                                                        & 30.43{\scriptsize $\pm$0.367}                                                        & 41.71                                                             \\
                                                                                                                                           & 61.91{\scriptsize $\pm$0.099}                                                       & 41.79{\scriptsize $\pm$0.206}                                                        & 73.42{\scriptsize $\pm$0.193}                                                        & 69.30{\scriptsize $\pm$0.184}                                                        & 45.80{\scriptsize $\pm$0.208}                                                        & 58.44                                                             \\ \hline
\multirow{2}{*}{WDC}                                                                                            & 44.44{\scriptsize $\pm$0.193}                                                        & 20.75{\scriptsize $\pm$0.212}                                                        & 58.49{\scriptsize $\pm$0.193}                                                        & 55.43{\scriptsize $\pm$0.308}                                                        & 30.52{\scriptsize $\pm$0.242}                                                        & 41.93                                                             \\
                                                                                                                                             & 62.27{\scriptsize $\pm$0.175}                                                       & 42.05{\scriptsize $\pm$0.198}                                                        & 73.58{\scriptsize $\pm$0.182}                                                        & 69.20{\scriptsize $\pm$0.203}                                                        & 45.87{\scriptsize $\pm$0.125}                                                        & 58.59                                                             \\ \hline

\multirow{2}{*}{\begin{tabular}[c]{@{}c@{}}Word-Adaptive \\ Domain Mixing\end{tabular}}                                     & 41.88{\scriptsize $\pm$0.240}                                                        & 19.84{\scriptsize $\pm$0.297}                                                       & 55.82{\scriptsize $\pm$0.594}                                                       & 52.88{\scriptsize $\pm$0.785}                                                       & 30.39{\scriptsize $\pm$0.141}                                                       & 40.16                                                             \\
                                                                                                                                             & 60.37{\scriptsize $\pm$0.113}                                                       & 41.02{\scriptsize $\pm$0.212}                                                       & 71.79{\scriptsize $\pm$0.290}                                                       & 67.62{\scriptsize $\pm$0.396}                                                        & 45.63{\scriptsize $\pm$0.113}                                                       & 57.29                                                             \\ \hline
                                                                                        
\multirow{2}{*}{Domain-Adapter}                                                                                                & 44.50{\scriptsize $\pm$0.342}                                                        & 20.37{\scriptsize $\pm$0.193}                                                        & 58.22{\scriptsize $\pm$0.169}                                                        & 56.00{\scriptsize $\pm$0.243}                                                        & 31.02{\scriptsize $\pm$0.334}                                                       & 42.02                                                             \\
                                                                                                                                             & 62.30{\scriptsize $\pm$0.248}                                                        & 41.65{\scriptsize $\pm$0.160}                                                        & 73.40{\scriptsize $\pm$0.066}                                                       & 69.54{\scriptsize $\pm$0.149}                                                        & 46.30{\scriptsize $\pm$0.306}                                                        & 58.64                                                             \\ \hline\hline
\multirow{2}{*}{\begin{tabular}[c]{@{}c@{}}Ours\\ (w/o $\mathcal{L_{\MI}}$)\end{tabular}}                                                           & 44.65{\scriptsize $\pm$0.318}                                                        & 20.43{\scriptsize $\pm$0.286}                                                        & 58.21{\scriptsize $\pm$0.692}                                                        & 55.38{\scriptsize $\pm$0.684}                                                        & 30.82{\scriptsize $\pm$0.498}                                                        & 41.90                                                             \\
                                                                                                                                             & 62.49{\scriptsize $\pm$0.221}                                                        & 41.77{\scriptsize $\pm$0.262}                                                        & 73.40{\scriptsize $\pm$0.416}                                                       & 69.28{\scriptsize $\pm$0.377}                                                       & 46.16{\scriptsize $\pm$0.414}                                                        & 58.62                                                             \\ \hline
\multirow{2}{*}{\begin{tabular}[c]{@{}c@{}}Ours\\ \end{tabular}}                                                            & \textbf{\begin{tabular}[c]{@{}c@{}}45.89{\scriptsize $\pm$0.215}\\ (+1.39)\end{tabular}} & \textbf{\begin{tabular}[c]{@{}c@{}}20.80{\scriptsize $\pm$0.298} \\ (+0.43)\end{tabular}} & \textbf{\begin{tabular}[c]{@{}c@{}}59.22{\scriptsize $\pm$0.306} \\ (+1.00)\end{tabular}} & \textbf{\begin{tabular}[c]{@{}c@{}}56.34{\scriptsize $\pm$0.238}\\ (+0.34)\end{tabular}} & \textbf{\begin{tabular}[c]{@{}c@{}}31.56{\scriptsize $\pm$0.218} \\ (+0.54)\end{tabular}} & \textbf{\begin{tabular}[c]{@{}c@{}}42.76 \\ (+0.74)\end{tabular}} \\
                                                                                                                                            & \textbf{\begin{tabular}[c]{@{}c@{}}63.19{\scriptsize $\pm$0.204} \\ (+0.89)\end{tabular}} & \textbf{\begin{tabular}[c]{@{}c@{}}42.05{\scriptsize $\pm$0.274} \\ (+0.39)\end{tabular}} & \textbf{\begin{tabular}[c]{@{}c@{}}74.02{\scriptsize $\pm$0.219} \\ (+0.62)\end{tabular}} & \textbf{\begin{tabular}[c]{@{}c@{}}69.94{\scriptsize $\pm$0.238} \\ (+0.40)\end{tabular}} & \textbf{\begin{tabular}[c]{@{}c@{}}46.46{\scriptsize $\pm$0.261} \\ (+0.16)\end{tabular}} & \textbf{\begin{tabular}[c]{@{}c@{}}59.13\\ (+0.49)\end{tabular}} \\
                                                                                        \bottomrule
\end{tabular}
}
\caption{\label{tab:main_result} Average and standard deviation of BLEU (upper line) and chrF (bottom line) from five random seed experiments. Bold indicates the best performance within a domain. Our model outperforms all baselines with significant margins ($p <0.05$).}
\end{center}
\end{table*}

Table~\ref{tab:main_result} presents sacreBLEU \cite{post-2018-call} and chrF \citep{popovic2015chrf} score from each model in all domains. 
For a fair comparison, we matched the number of parameters for all models.
Baseline results following its original implementation with different parameter size are provided in Appendix~\ref{sec:additional_base_line_results}.
Interestingly, Mixed performs on par with Domain-Tag and outperforms Word-Adaptive Domain Mixing, suggesting that not all multi-domain NMT methods are effective.
Although adapter-based models (\textit{i.e.}, Ours (w/o $\mathcal{L_\MI}$) and Domain-Adapter) outperform Mixed, the performance increase is still marginal.
Our model has gained 1.15 BLEU improvement over Mixed. 
It also outperforms all baselines with statistically significant difference.

As an ablation study of our MI objective, we conduct experiments without $\mathcal{L_\text{MI}}$ to prove its effectiveness. 
The result confirms that $\mathcal{L}_\text{MI}$ encouraged the model to learn domain specific knowledge leading to refined translation.  

\subsection{Mutual Information Distribution}

\begin{figure}
    \centering
    \includegraphics[width=\columnwidth]{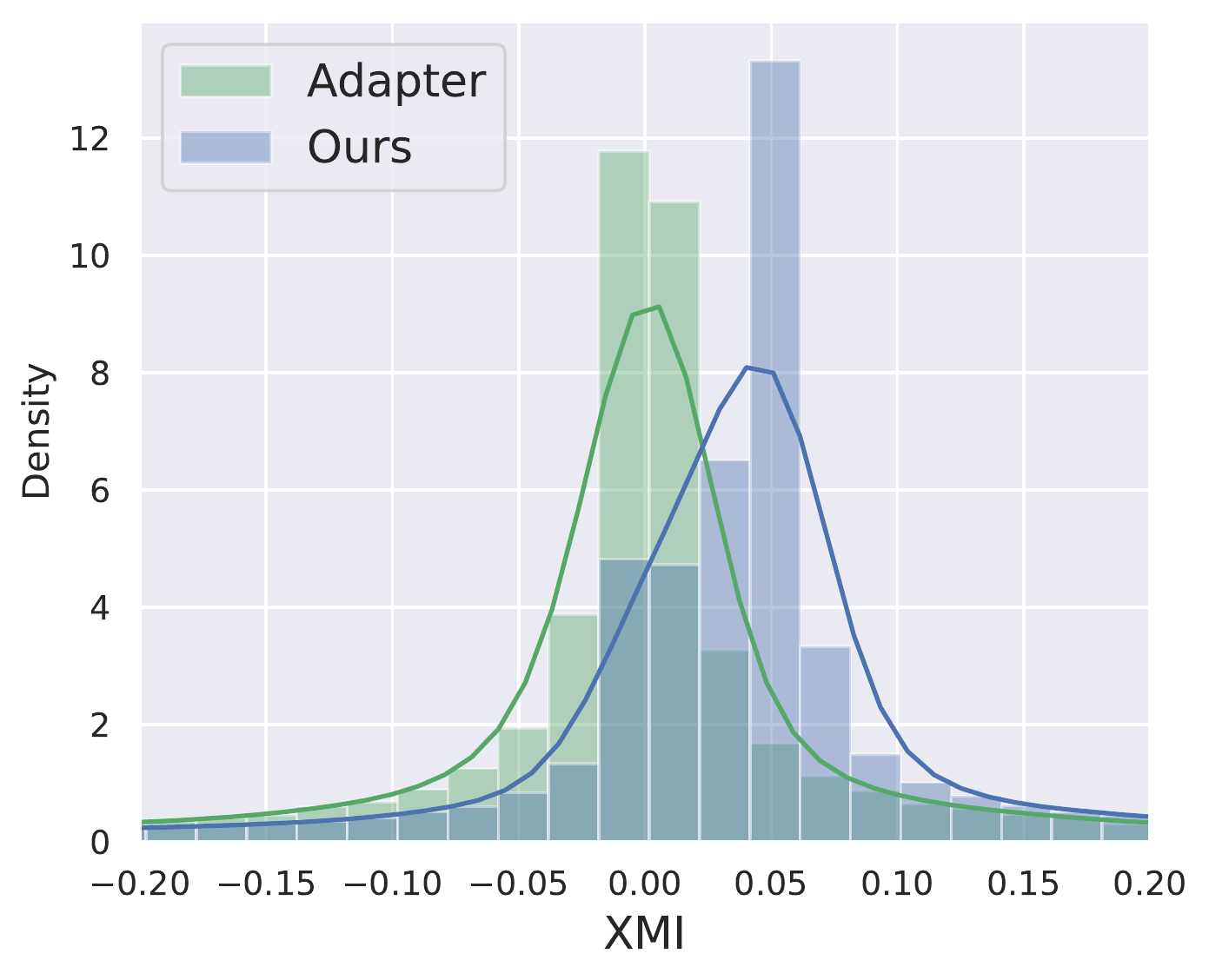}
    \caption{XMI distribution in Law. X-axis is XMI and Y-axis is the density. Green is Domain-Adapter and blue is our model. Our model has more high XMI values.}
    \label{fig:koran_dist}
\end{figure}

We visualize $\XMI(i)$ in Eq.~(\ref{delta_i}) to verify that our proposed loss penalizes low XMI.
Figure~\ref{fig:koran_dist} is the histogram of $\XMI(i)$ from the test samples in Law. 
Other domain distributions are in Appendix~\ref{sec:mutual_information_distribution}. 
We use Domain-Adapter for comparison since it performs the best among the baselines. 
For $p_G$, we use the output probability of Mixed for both cases.
From the distributions, our method indeed penalizes low XMI and encourages the model to have high XMI in all domains.

\subsection{Translation Performance for Domain Specialized Sentences}
\begin{table}[]
\begin{center}
\resizebox{\linewidth}{!}{
\begin{tabular}{c|cccc|c}
\toprule
          & -Q1 & Q1-Q2 & Q2-Q3 & Q3-Q4 & Average\\ \hline
IT        & 2.26   & 1.14      & 1.12      & 1.27  & 1.39  \\
Koran     & 0.97   & 0.65      & 0.49      & 0.41  & 0.43     \\
Law       & 0.75   & 0.73      & 1.41      & 1.17  & 1.00     \\
Medical   & 0.25   & 0.08      & 0.60      & 0.40  & 0.34      \\
Subtitles & 0.36   & 0.83      & 0.83      & 2.10  & 0.54    \\
\bottomrule
\end{tabular}
}
\caption{\label{translation_performance_ratios} BLEU improvements in each quartiles. Q1, Q2, Q3 and Q4 represents 25\%, 50\%, 75\% 100\% respectively. The higher the quartile, the more domain specific the samples. }
\end{center}
\end{table}
Since the ultimate goal is to specialize multi-domain NMT, we calculate BLEU score improvement according to the domain specificity. 
We extract top 1\% TF-IDF words in train source sentences in each domain (examples are in Appendix~\ref{sec:TF-IDF words}) and consider them as domain-specific keywords.
We assume that the more these keywords are included in the source sentence, the more domain specialized the sample is.
 We divide the test set into quartiles based on the number of the keywords the source sentence contains.

Table~\ref{translation_performance_ratios} reports BLEU score improvement compared to Domain-Adapter in each quartile along with averaged performance increases.
In Law, Medical and Subtitles, BLEU score improvement increases as quartile gets higher.
Furthermore, the improvements in Q2-Q3 and Q3-Q4 are larger than the averaged improvement score (\textit{i.e.}, Average column).
However, in IT and Koran, -Q1 has the largest performance increases.
We conjecture the reason is that in both domains, the number of top TF-IDF words include in higher quartiles is fewer than the other domains.
This weak distinction among quartiles in IT and Koran can be the root cause of marginal performance improvement.
Details on number of captured keywords are in Appendix~\ref{sec:TF-IDF words}.

\subsection{Experiment Results on Korean-English}
\begin{table}[]
\begin{center}
\resizebox{\linewidth}{!}{
\begin{tabular}{c|ccc|c}
\toprule
                                & Finance                                                                & Ordinance                                                              & Tech                                                                   & Average                                                          \\ \hline
\multirow{2}{*}{Mixed}      & 52.50{\scriptsize $\pm$0.220}                                                            & 56.65{\scriptsize $\pm$0.100}                                                            & 66.00{\scriptsize $\pm$0.242}                                                            & 58.38                                                            \\
                                & 72.64{\scriptsize $\pm$0.105}                                                            & 75.36{\scriptsize $\pm$0.091}                                                            & 81.60{\scriptsize $\pm$0.121}                                                            & 76.53                                                            \\ \hline
\multirow{2}{*}{Domain-Tag} & 52.71{\scriptsize $\pm$0.231}                                                            & 56.60{\scriptsize $\pm$0.115}                                                            & 66.03{\scriptsize $\pm$0.360}                                                            & 58.45                                                            \\
                                & 72.77{\scriptsize $\pm$0.175}                                                            & 75.38{\scriptsize $\pm$0.058}                                                            & 81.64{\scriptsize $\pm$0.185}                                                            & 76.60                                                            \\ \hline
\multirow{2}{*}{WDC} & 52.75{\scriptsize $\pm$0.136}                                                            & 56.56{\scriptsize $\pm$0.124}                                                            & 65.93{\scriptsize $\pm$0.214}                                                            & 58.41                                                            \\
                                & 72.78{\scriptsize $\pm$0.135}                                                            & 75.34{\scriptsize $\pm$0.053}                                                            & 81.53{\scriptsize $\pm$0.099}                                                            & 76.55                                                            \\ \hline
\multirow{2}{*}{Domain-Adapter} & 53.13{\scriptsize $\pm$0.186}                                                            & 56.97{\scriptsize $\pm$0.129}                                                            & 66.25{\scriptsize $\pm$0.103}                                                            & 58.78                                                            \\
                                & 72.98{\scriptsize $\pm$0.170}                                                            & 75.48{\scriptsize $\pm$0.066}                                                            & 81.76{\scriptsize $\pm$0.079}                                                            & 76.74                                                            \\ \hline \hline
\multirow{2}{*}{Ours}           & \textbf{\begin{tabular}[c]{@{}c@{}}53.87{\scriptsize $\pm$0.188}\\ (+0.74)\end{tabular}} & \textbf{\begin{tabular}[c]{@{}c@{}}57.47{\scriptsize $\pm$0.086}\\ (+0.50)\end{tabular}} & \textbf{\begin{tabular}[c]{@{}c@{}}66.66{\scriptsize $\pm$0.191}\\ (+0.41)\end{tabular}} & \textbf{\begin{tabular}[c]{@{}c@{}}59.33\\ (+0.55)\end{tabular}} \\
                                & \textbf{\begin{tabular}[c]{@{}c@{}}73.41{\scriptsize $\pm$0.162}\\ (+0.43)\end{tabular}} & \textbf{\begin{tabular}[c]{@{}c@{}}75.81{\scriptsize $\pm$0.033}\\ (+0.33)\end{tabular}} & \textbf{\begin{tabular}[c]{@{}c@{}}81.99{\scriptsize $\pm$0153}\\ (+0.23)\end{tabular}}  & \textbf{\begin{tabular}[c]{@{}c@{}}77.07\\ (+0.33)\end{tabular}} \\
\bottomrule
\end{tabular}
}
\caption{\label{tab:Ko-En_experiment} Average and standard deviation of BLEU (upper line) and chrF (bottom line) from five random seeds on Ko-En dataset. Bold indicates the best performance within a domain.}
\end{center}
\end{table}
To verify the effectiveness of our proposed method in different language, we additionally conducted experiment on Korean-English dataset which has approximately 1M samples with three domains: Finance, Ordinance and Tech. 
The dataset is obtained from AIhub\footnote{https://aihub.or.kr/} which is publicly available. 
Model configuration is identical with the main experiment on OPUS. 
More experimental details are in Appendix~\ref{sec:Experiment Details}.

Table~\ref{tab:Ko-En_experiment} demonstrates the results from the major baselines and our model, where we select top-4 baselines in the experiment on OPUS (Table~\ref{tab:main_result}).
Our model achieves  the best performance in all three domains, outperforming Domain-Adpater by 0.55 BLEU score on average.
This result confirms that our proposed method can be further extended to other languages.

\subsection{Samples with MI Visualization}
\begin{figure}
    \centering
    \includegraphics[width=\columnwidth]{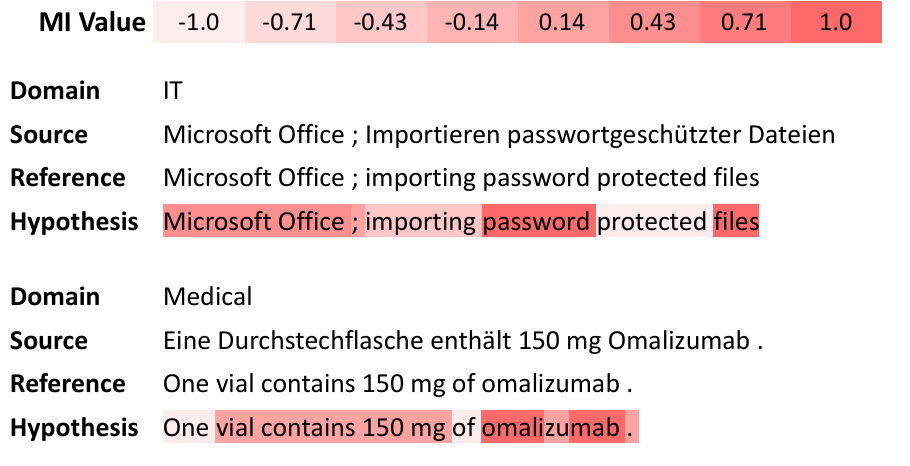}
    \caption{Example visualizations with MI values from IT and Medical. The more intense the red, the more higher MI value.}
    \label{fig:main_example_visualization}
\end{figure}
Figure~\ref{fig:main_example_visualization} demonstrates test set outputs with MI values generated by our model.
Color intensity is correlated with MI value; the more intense the red, the more higher MI value.
Note that the model has high MI especially when generate domain-specific words (\textit{e.g.}, `password' in IT and `omalizumab' in Medical).
This result is analogous to our motivation in that high MI value encourages model to translate domain peculiar terms.
More samples are provided in Appendix~\ref{app:sample_with_MI_Visualization}.

\subsection{Computation Cost for Training}
\begin{table}[bt!]
\begin{center}
\resizebox{\linewidth}{!}{
\begin{tabular}{c|cc}
\toprule
                   & Domain-Adapter & Ours    \\ \hline
Number of Iterations ($\downarrow$)    & 51.7K           & 48.5K   \\
Peak Memory (GB) ($\downarrow$)   & 26.79          & 27.09  \\
Words per Second ($\uparrow$)  & 22.9K        & 22K \\
Updates per Second ($\uparrow$) & 0.4            & 0.38    \\
\bottomrule
\end{tabular}
}
\caption{\label{tab:computation_cost} Comparison of training computation cost between Domain-Adapter and Ours. The values are averaged across five seed experiments.}
\end{center}
\end{table}
We compare computation cost between Domain-Adapter and our proposed model. 
Domain-Adapter was chosen because it shares the same model architecture with ours. 
Table~\ref{tab:computation_cost} provides four computation cost during training: Number of Iterations until converge, Peak Memory, Words per Second, and Updates per Second. 
Our model requires fewer number of iterations needed to be trained (3.2K difference on average across five seeds). 
Our model has slightly higher peak memory (0.3GB, 0.94\%) than Domain-Adapter, however, we believe this is acceptable when considering the performance improvement. 
Furthermore, there was not much difference in words per second and updates per second during training.
\section{Conclusion}
We build a specialized multi-domain NMT by adding MI-based loss.
We reinterpret domain-specific knowledge from MI perspective and promote a model to explore domain knowledge by penalizing low MI.
Our results prove that the proposed method is effective in increasing overall MI.
\section*{Limitations}

Although many previous multi-domain NMT studies regard the source of the given sentence as its domain, equating domain and corpora is a naive approach and can partially represent the data. \citep{aharoni-goldberg-2020-unsupervised} 
For instance, some sentences may incorporate multiple domain characteristics or can be better translated under different domain other than its source domain. \citep{currey-etal-2020-distilling, pham2021revisiting}
This problem is not limited to our work but is applicable to other previous multi-domain NMT studies. 
Establishing a more proper definition of domain is a future work and a critical challenge in multi-domain NMT.






\section*{Acknowledgement}
The authors would like to thank all members in Papago Team, NAVER Corporation for their valuable comments.
Also, we sincerely thank Jin-Hwa Kim at NAVER AI Lab for the insightful feedback.
This work was partly supported by Institute of Information \& Communications Technology Planning \& Evaluation (IITP) grant (No.2019-0-00075, Artificial Intelligence Graduate School Program(KAIST)), and National Research Foundation of Korea (NRF) grant (NRF-2020H1D3A2A03100945) funded by the Korea government (MSIT).
\bibliography{EMNLP2022/anthology,EMNLP2022/custom}
\bibliographystyle{EMNLP2022/acl_natbib}

\newpage
\appendix
\section{Experiment Settings}
\label{sec:Experiment Details}
\subsection{Dataset}
For preprocessing, we conducted tokenization and normalize-punctuation by using Moses \citep{koehn2007moses} pipeline. 
We additionally eliminated samples with (i) sequences shorter than one subword-token, (ii) sequences longer than 250 subword-tokens, (iii) severe length imbalance between the language pair (top, bottom 5\% for each domain) for both De-En and Ko-En.
Table~\ref{deen number of samples} and~\ref{koen number of samples} show the final number of samples. 

\begin{table}[hbt!]
\begin{center}

\begin{tabular}{cccc}
\toprule
De-En     & Train   & Dev   & Test  \\ \hline
IT        & 211,374 & 1,888 & 2,000 \\
Koran     & 16,952  & 1,872 & 2,000 \\
Law       & 434,555 & 1,861 & 2,000 \\
Medical   & 233,167 & 1,873 & 2,000 \\
Subtitles & 470,611 & 1,899 & 2,000 \\
\bottomrule
\end{tabular}
\caption{\label{deen number of samples} Number of samples in De-En}
\end{center}
\end{table}
\begin{table}[hbt!]
\begin{center}
\begin{tabular}{cccc}
\toprule
Ko-En     & Train   & Dev   & Test  \\ \hline
Finance   & 156,569 & 9,510 & 5,000 \\
Ordinance & 79,802  & 9,335 & 5,000 \\
Tech      & 711,885 & 9,251 & 5,000 \\
\bottomrule
\end{tabular}
\caption{\label{koen number of samples} Number of samples in Ko-En}
\end{center}
\end{table}

\subsection{Experiment Details}

For De-En, we use a joint BPE vocabulary \citep{sennrich-etal-2016-neural} learned with 32k merge operations. For Ko-En, we train BPE vocabulary with 32k size separately for each language since Korean and English do not share characters.
Remainig experiment settings are identical for both language pairs.
Our experiments are conducted under open-source fairseq\footnote{\url{https://github.com/facebookresearch/fairseq}} \citep{ott2019fairseq} framework.
We built upon Transformer model \citep{vaswani2017attention} which has 6 encoder and decoder layers with embedding dimension of 512, feed-forward dimension of 2048, and attention heads of 8.
Parameters of encoder embedding, decoder embedding and decoder last layer are shared.
We also utilize the same sinusoidal positional embedding following the original work.
We fix dropout to 0.1 and used ReLU \citep{agarap2018deep} as an activation function.
Following \citet{bapna2019simple}, the domain and general adapters ($\phi_1, \cdots \phi_N, \phi_G$) are inserted after feed-forward layer following the multi-head attention.
%
Note that our training differs in that we jointly train all parameters from scratch including adapters on all domains.

The bottleneck size of the adapters is 256.
Adapters are initialized from zero-mean Gaussian with standard deviation $10^{-2}$ which proven to be most effective in the proposed work.
We searched the best combination of $\lambda_1$ and $\lambda_2$ by grid search ranging from 0.5 to 1.0.
Then, we set $\lambda_1$ and $\lambda_2$ both to 1.
All experiments are trained with label-smoothing cross-entropy loss with smoothing parameter of 0.1.
All experiments are conducted using 8 NVIDIA V100 GPU.

In all experiments, a model is trained until early stopping with patience of 10 based on BLEU.
We use Adam \citep{kingma2014adam} optimizer with an initial learning rate of 5$\cdot e^{-4}$, where the learning rate is searced within the range of 0.001 to 0.0001. 
The tokens per batch is 8192 in all experiments.
We compute sacreBLEU score~\footnote{sacreBLEU signature: BLEU+c.mixed+l.de-en+\#.1+s.

exp+tok.13a+v.2.0.0} from outputs using beam search with a beam size of 5.

\subsection{Baselines}
In this section, we provide a detailed explanation on each baseline.
Mixed and Domain-Tag employ identical model architecture with our model excluding adapters.
Mixed does not utilize domain information and treat all samples are from identical distribution.
On the other hand, Domain-Tag distinguishes domain by adding domain tag in front of the input sentence.
We enlarged Mixed and Domain-Tag to Mixed-Big and Domain-Tag-big by increasing encoder and decoder embedding dimension to 608.

For Word-Adaptive Domain Mixing, we borrowed publicly available code and applied on our dataset. 
We applied word-adaptive training in both encoder and decoder because it had the best performance in the original paper. Other configurations are the same with ours.

Domain-Adapter has the domain-specific adapters $\phi_1, \cdots, \phi_N$, and the input sentence passes through only its domain adapter. 
Note that there are two major differences between Domain-Adapter and ours. 
First, Domain-Adapter does not need general adapter $\phi_G$ since it does not calculate $p_G(Y|X)$.
Second, a source sentence only passes through the model once only through its domain adapter $\phi_d$.
Similar to our method, all the parameters including adapters are jointly trained from scratch.

\section{Full Derivation of Domain-Aware Mutual Information}
\label{sec:DAMI derivation}

Below is the full derivation of Domain-Aware Mutual Information.

\begin{align*}
    MI(D;Y|X) &= \mathbb{E}_{D,X,Y} \bigg[ \log \frac{p(D, Y|X)}{p(D|X) \cdot p(Y|X)} \bigg] \\
    &= \mathbb{E}_{D,X,Y} \bigg[ \log \frac{p(D|Y,X) \cdot \cancel{p(Y|X)}}{p(D|X) \cdot \cancel{p(Y|X)}} \bigg] \\
    &= \mathbb{E}_{D,X,Y} \bigg[ \log \frac{p(X,Y,D) \cdot p(X)}{p(X,Y) \cdot p(X,D)} \bigg] \\
    &= \mathbb{E}_{D,X,Y} \bigg[ \log \frac{p(Y|X,D)}{p(Y|X)} \bigg]
\end{align*}

The proof from the first to the second line is provided below.
\begin{align*}
    P(X,Y,D) &= P(D|Y,X) \cdot P(Y|X) \cdot P(X) \\
    \Rightarrow P(D,Y|X) &= P(D|Y,X) \cdot P(Y|X)
\end{align*}

\section{Additional Baseline Results}
\label{sec:additional_base_line_results}
\begin{table*}[]
\begin{center}
\resizebox{\linewidth}{!}{
\begin{tabular}{c|c|ccccc|c}
\toprule
                                                                                        & \begin{tabular}[c]{@{}c@{}}\# of \\ Parameters\end{tabular} & IT                                                                  & Koran                                                               & Law                                                                 & Medical                                                             & Subtitles                                                           & Average                                                           \\ \hline
\multirow{2}{*}{Mixed-Small}                                                                  & \multirow{2}{*}{60M}                                        & 43.64{\scriptsize $\pm$0.253}                                                       & 20.74{\scriptsize $\pm$0.155}                                                       & 57.47{\scriptsize $\pm$0.376}                                                     & 54.88{\scriptsize $\pm$0.553}                                                     & 30.58{\scriptsize $\pm$0.273}                                                    & 41.46                                                             \\
                                                                                        &                                                             & 61.96{\scriptsize $\pm$0.204}                                                       & 42.01{\scriptsize $\pm$0.159}                                                       & 72.95{\scriptsize $\pm$0.275}                                                       & 68.98{\scriptsize $\pm$0.362}                                                       & 46.06{\scriptsize $\pm$0.293}                                                       & 58.39                                                             \\ \hline
                                                                                        \multirow{2}{*}{Mixed}    
                                                                                        & \multirow{2}{*}{76M}   & 43.87{\scriptsize $\pm$0.505}                                                        & 20.31{\scriptsize $\pm$0.371}                                                        & 58.33{\scriptsize $\pm$0.474}                                                        & 55.19{\scriptsize $\pm$0.737}                                                        & 30.36{\scriptsize $\pm$0.424}                                                       & 41.61                                                             \\
                                                                                                                                                   & & 62.00{\scriptsize $\pm$0.403}                                                       & 41.75{\scriptsize $\pm$0.343}                                                       & 73.41{\scriptsize $\pm$0.303}                                                       & 69.14{\scriptsize $\pm$0.346}                                                       & 45.73{\scriptsize $\pm$0.424}                                                       & 58.40                                                             \\ \hline
\multirow{2}{*}{Domain-Tag-Small}                                                             & \multirow{2}{*}{60M}                                        & 44.00{\scriptsize $\pm$0.409}                                                        & 20.39{\scriptsize $\pm$0.251}                                                        & 57.80{\scriptsize $\pm$0.206}                                                        & 54.91{\scriptsize $\pm$0.197}                                                       & 31.10{\scriptsize $\pm$0.316}                                                       & 41.64                                                             \\
                                                                                        &                                                             & 62.09{\scriptsize $\pm$0.242}                                                       & 41.71{\scriptsize $\pm$0.141}                                                       & 73.14{\scriptsize $\pm$0.183}                                                       & 68.97{\scriptsize $\pm$0.056}                                                       & 46.34{\scriptsize $\pm$0.208}                                                       & 58.45                                                             \\ \hline
                                                                                        \multirow{2}{*}{Domain-Tag}                                                                      & \multirow{2}{*}{76M}                               & 44.29{\scriptsize $\pm$0.142}                                                        & 20.44{\scriptsize $\pm$0.236}                                                        & 58.47{\scriptsize $\pm$0.275}                                                        & 55.39{\scriptsize $\pm$0.288}                                                        & 30.61{\scriptsize $\pm$0.220}                                                        & 41.84                                                             \\
                                                                                                                &                             & 62.30{\scriptsize $\pm$0.111}                                                       & 41.75{\scriptsize $\pm$0.203}                                                        & 73.56{\scriptsize $\pm$0.190}                                                        & 69.28{\scriptsize $\pm$0.160}                                                        & 45.99{\scriptsize $\pm$0.268}                                                        & 58.58                                                             \\ \hline
                                                                                                                                             \multirow{2}{*}{MTL}      & \multirow{2}{*}{76M}                                                                                   & 44.00{\scriptsize $\pm$0.298}                                                        & 20.40{\scriptsize $\pm$0.198}                                                        & 58.27{\scriptsize $\pm$0.327}                                                        & 55.24{\scriptsize $\pm$0.564}                                                        & 30.52{\scriptsize $\pm$0.478}                                                        & 41.69                                                             \\
                                                                                         &                                                  & 62.11{\scriptsize $\pm$0.169}                                                       & 41.78{\scriptsize $\pm$0.174}                                                        & 73.42{\scriptsize $\pm$0.197}                                                        & 69.16{\scriptsize $\pm$0.235}                                                        & 45.87{\scriptsize $\pm$0.316}                                                        & 58.47                                                             \\ \hline
\multirow{2}{*}{AdvL}                        & \multirow{2}{*}{76M}                                                                            & 43.86{\scriptsize $\pm$0.167}                                                        & 20.33{\scriptsize $\pm$0.275}                                                        & 58.40{\scriptsize $\pm$0.195}                                                        & 55.56{\scriptsize $\pm$0.245}                                                        & 30.43{\scriptsize $\pm$0.367}                                                        & 41.71                                                             \\
                                                                                                                                           & & 61.91{\scriptsize $\pm$0.099}                                                       & 41.79{\scriptsize $\pm$0.206}                                                        & 73.42{\scriptsize $\pm$0.193}                                                        & 69.30{\scriptsize $\pm$0.184}                                                        & 45.80{\scriptsize $\pm$0.208}                                                        & 58.44                                                             \\ \hline
\multirow{2}{*}{WDC}                    & \multirow{2}{*}{76M}                                                                        & 44.44{\scriptsize $\pm$0.193}                                                        & 20.75{\scriptsize $\pm$0.212}                                                        & 58.49{\scriptsize $\pm$0.193}                                                        & 55.43{\scriptsize $\pm$0.308}                                                        & 30.52{\scriptsize $\pm$0.242}                                                        & 41.93                                                             \\
                                                                                                                                   &          & 62.27{\scriptsize $\pm$0.175}                                                       & 42.05{\scriptsize $\pm$0.198}                                                        & 73.58{\scriptsize $\pm$0.182}                                                        & 69.20{\scriptsize $\pm$0.203}                                                        & 45.87{\scriptsize $\pm$0.125}                                                        & 58.59                                                             \\ \hline
                                                                                                                                             
\multirow{2}{*}{\begin{tabular}[c]{@{}c@{}}Word-Adaptive \\ Domain Mixing Big\end{tabular}} & \multirow{2}{*}{218M}                                       & 44.08{\scriptsize $\pm$0.561}                                                        & 20.34{\scriptsize $\pm$0.257}                                                       & 59.63{\scriptsize $\pm$0.308}                                                       & 56.81{\scriptsize $\pm$0.386}                                                       & 29.34{\scriptsize $\pm$0.488}                                                       & 42.04                                                             \\
                                                                                        &                                                             & 61.93{\scriptsize $\pm$0.273}                                                       & 41.23{\scriptsize $\pm$0.334}                                                       & 74.26{\scriptsize $\pm$0.196}                                                       & 69.96{\scriptsize $\pm$0.245}                                                        & 44.62{\scriptsize $\pm$0.460}                                                       & 58.40                                                             \\ \hline
                                                                                        \multirow{2}{*}{\begin{tabular}[c]{@{}c@{}}Word-Adaptive \\ Domain Mixing\end{tabular}}        & \multirow{2}{*}{76M}                             & 41.88{\scriptsize $\pm$0.240}                                                        & 19.84{\scriptsize $\pm$0.297}                                                       & 55.82{\scriptsize $\pm$0.594}                                                       & 52.88{\scriptsize $\pm$0.785}                                                       & 30.39{\scriptsize $\pm$0.141}                                                       & 40.16                                                             \\
                                                                                                                                 &            & 60.37{\scriptsize $\pm$0.113}                                                       & 41.02{\scriptsize $\pm$0.212}                                                       & 71.79{\scriptsize $\pm$0.290}                                                       & 67.62{\scriptsize $\pm$0.396}                                                        & 45.63{\scriptsize $\pm$0.113}                                                       & 57.29                                                             \\ \hline
                                                                                        
\multirow{2}{*}{Domain-Adapter}              & \multirow{2}{*}{76M}                                                                                  & 44.50{\scriptsize $\pm$0.342}                                                        & 20.37{\scriptsize $\pm$0.193}                                                        & 58.22{\scriptsize $\pm$0.169}                                                        & 56.00{\scriptsize $\pm$0.243}                                                        & 31.02{\scriptsize $\pm$0.334}                                                       & 42.02                                                             \\
                                                                                                                               &              & 62.30{\scriptsize $\pm$0.248}                                                        & 41.65{\scriptsize $\pm$0.160}                                                        & 73.40{\scriptsize $\pm$0.066}                                                       & 69.54{\scriptsize $\pm$0.149}                                                        & 46.30{\scriptsize $\pm$0.306}                                                        & 58.64
                                                                                                                                             \\ \hline \hline
                                                                                        \multirow{2}{*}{\begin{tabular}[c]{@{}c@{}}Ours\\ (w/o $\mathcal{L_{\MI}}$)\end{tabular}}     & \multirow{2}{*}{76M}                                                      & 44.65{\scriptsize $\pm$0.318}                                                        & 20.43{\scriptsize $\pm$0.286}                                                        & 58.21{\scriptsize $\pm$0.692}                                                        & 55.38{\scriptsize $\pm$0.684}                                                        & 30.82{\scriptsize $\pm$0.498}                                                        & 41.90                                                             \\
                                                                                       &                                                      & 62.49{\scriptsize $\pm$0.221}                                                        & 41.77{\scriptsize $\pm$0.262}                                                        & 73.40{\scriptsize $\pm$0.416}                                                       & 69.28{\scriptsize $\pm$0.377}                                                       & 46.16{\scriptsize $\pm$0.414}                                                        & 58.62                                                             \\ \hline
\multirow{2}{*}{\begin{tabular}[c]{@{}c@{}}Ours\\ \end{tabular}} & \multirow{2}{*}{76M}                                                            & \textbf{\begin{tabular}[c]{@{}c@{}}45.89{\scriptsize $\pm$0.215}\\ (+1.39)\end{tabular}} & \textbf{\begin{tabular}[c]{@{}c@{}}20.80{\scriptsize $\pm$0.298} \\ (+0.43)\end{tabular}} & \textbf{\begin{tabular}[c]{@{}c@{}}59.22{\scriptsize $\pm$0.306} \\ (+1.00)\end{tabular}} & \textbf{\begin{tabular}[c]{@{}c@{}}56.34{\scriptsize $\pm$0.238}\\ (+0.34)\end{tabular}} & \textbf{\begin{tabular}[c]{@{}c@{}}31.56{\scriptsize $\pm$0.218} \\ (+0.54)\end{tabular}} & \textbf{\begin{tabular}[c]{@{}c@{}}42.76 \\ (+0.74)\end{tabular}} \\
                                                                                  &                                                          & \textbf{\begin{tabular}[c]{@{}c@{}}63.19{\scriptsize $\pm$0.204} \\ (+0.89)\end{tabular}} & \textbf{\begin{tabular}[c]{@{}c@{}}42.05{\scriptsize $\pm$0.274} \\ (+0.39)\end{tabular}} & \textbf{\begin{tabular}[c]{@{}c@{}}74.02{\scriptsize $\pm$0.219} \\ (+0.62)\end{tabular}} & \textbf{\begin{tabular}[c]{@{}c@{}}69.94{\scriptsize $\pm$0.238} \\ (+0.40)\end{tabular}} & \textbf{\begin{tabular}[c]{@{}c@{}}46.46{\scriptsize $\pm$0.261} \\ (+0.16)\end{tabular}} & \textbf{\begin{tabular}[c]{@{}c@{}}59.13\\ (+0.49)\end{tabular}} \\
                               \bottomrule
\end{tabular}
}
\caption{\label{tab:app_main_result} Average and standard deviation of BLEU (upper line) and chrF (bottom line) from baselines with different parameter size and our model.}
\end{center}
\end{table*}
Table~\ref{tab:app_main_result} shows sacreBLEU \cite{post-2018-call} and chrF \citep{popovic2015chrf} score from baseline models (Mixed-Small, Domain-Tag-Small, Word-Adaptive Domain Mixing Big) following its original implementation with different number of parameters. Note that Mixed and Domain-Tag underperform the models in the main experiment, and Word-Adaptive Domain Mixing becomes effective when enlarge model size.

\section{MI Histogram}
\label{sec:mutual_information_distribution}

\begin{figure*}
    \centering
    \includegraphics[width=\textwidth]{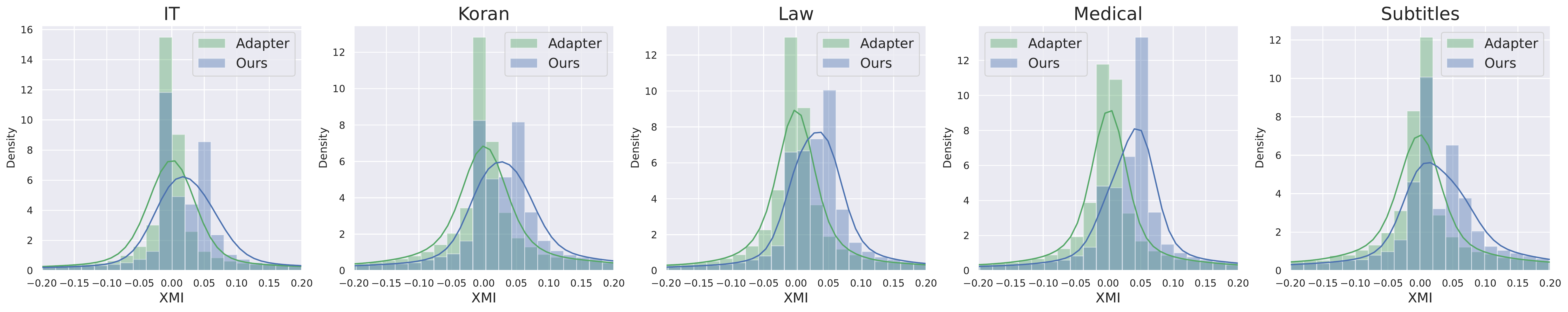}
    \caption{XMI distributions from all domains. X-axis is XMI and Y-axis is the density. Adpater is colored in green and our model is in blue. Our XMI distribution has more higher values than baseline.}
    \label{fig:5domains_dist}
\end{figure*}

XMI histograms from all doamins are in Fig.~\ref{fig:5domains_dist}. Adpater (baseline) is colored in green and our model is in blue. From the plot, we can verify that in all domains, our model outputs higher XMI compared to the baseline. 

\section{Details on Translation Performance for Domain Specialized Experiment}
\label{sec:TF-IDF words}
\begin{table*}[]
\begin{tabularx}{\textwidth}{c|X}
\toprule
\multicolumn{1}{l|}{} & Examples of TF-IDF Keywords                                                                                                                                                                                   \\ \hline
IT                    & übertragungsrate (transfer rate), zwischensumme (temporally save), YouTube, übertragungsfortschritt (transfer progress), speichergröße (memory size), yahoomail, zusammensetzungswerkzeug (composition tool), übersichtsmodus (overview mode), webserver, zwischengespeichert (cached), übersetzungsprogramm (translation game), zwischenablagename (clipboard name) \\ \hline
Koran                 & übertreter (transgressor), zwingherr (tyrant), widmest (dedicate), unterwürfig (submissive), unterworfen (subjected), städte (cities), sterben (die), schutzherr (patron), religion, muslime (muslims)                                                                                                  \\ \hline
Law                   & überwachungszollstelle (supervising customs office), änderungsverfahren (change procedure), zustellungsmängel (delivery defects), wirtschaftsjahre (fiscal years), überstunden (overtime), zuschusssatz (subsidy rate), widerklänge (echoes), übernahmeprotokoll (takeover protocol), zulassungsvoraussetzungen (admission requirements), verwaltungskommission (administrative commission), tarife (rates)     \\ \hline
Medical               & überlebenswahrscheinlichkeit (probability of survival), verletzung (injury), tagesgesamtdosis (total daily dose), überempfindlichkeitsreaktionen (hypersensitivity reactions), zäpfchen (suppository), tremor, zytotoxisch (cytotoxic), urin (urine), Schatzungen (estimates), wirkstoffmatrix (active ingredient matrix), vorsichtsmaßnahmen (precautions)                             \\ \hline
Subtitles             & übungen (exercises) , wähle (choose), übersehen (overlook), öffentlichkeitsarbeit (public relation), äußern (to express), ärger (trouble), ältester (oldest), zähflüssig (viscous), zwischenmahlzeiten (snacks), wäsche (laundry), werbepause (commercial break)        \\
\bottomrule
\end{tabularx}
\caption{\label{tab:tf-idf word example} Examples of TF-IDF extracted words of the source language (i.e., German). We randomly sampled 11 words from each domain among the top 1\% keywords. We also write its meaning in english words in the bracket.}
\end{table*}
Examples of extracted TF-IDF keywords are in Table~\ref{tab:tf-idf word example}. 
We removed stop words and conducted lemmatization before extracting keywords.
As expected, chosen words are correlated to its domain are chosen.

\begin{table}[hbt!]
\begin{center}
\resizebox{\linewidth}{!}{
\begin{tabular}{cccccc}
\toprule
      & IT   & Koran & Law   & Medical & Subtitles \\ \hline
-Q1   & 0    & 0     & 1.91  & 0.96    & 1.70      \\ 
Q1-Q2 & 1    & 1     & 5.07  & 4.05    & 3.57      \\ 
Q2-Q3 & 2    & 2     & 8.36  & 6.83    & 5.42      \\ 
Q3-Q4 & 4.10 & 3.42  & 13.21 & 11.50   & 8.11   \\
\bottomrule
\end{tabular}
}
\caption{\label{tab:averaged_capture_words} Averaged number of captured top TF-IDF keywords in each quartile.}

\end{center}
\end{table}
Averaged number of captured keywords in each quartile is presented in Table~\ref{tab:averaged_capture_words}.
Compared to Law, Medical and Subtitles where a clear distinction among quartiles by the averaged number of keywords exists, IT and Koran have a minimal change indicating a weak distinction.

\section{Samples with MI Visualization}
\label{app:sample_with_MI_Visualization}
\begin{figure*}[ht!]
    \centering
    \includegraphics[width=0.8\textwidth]{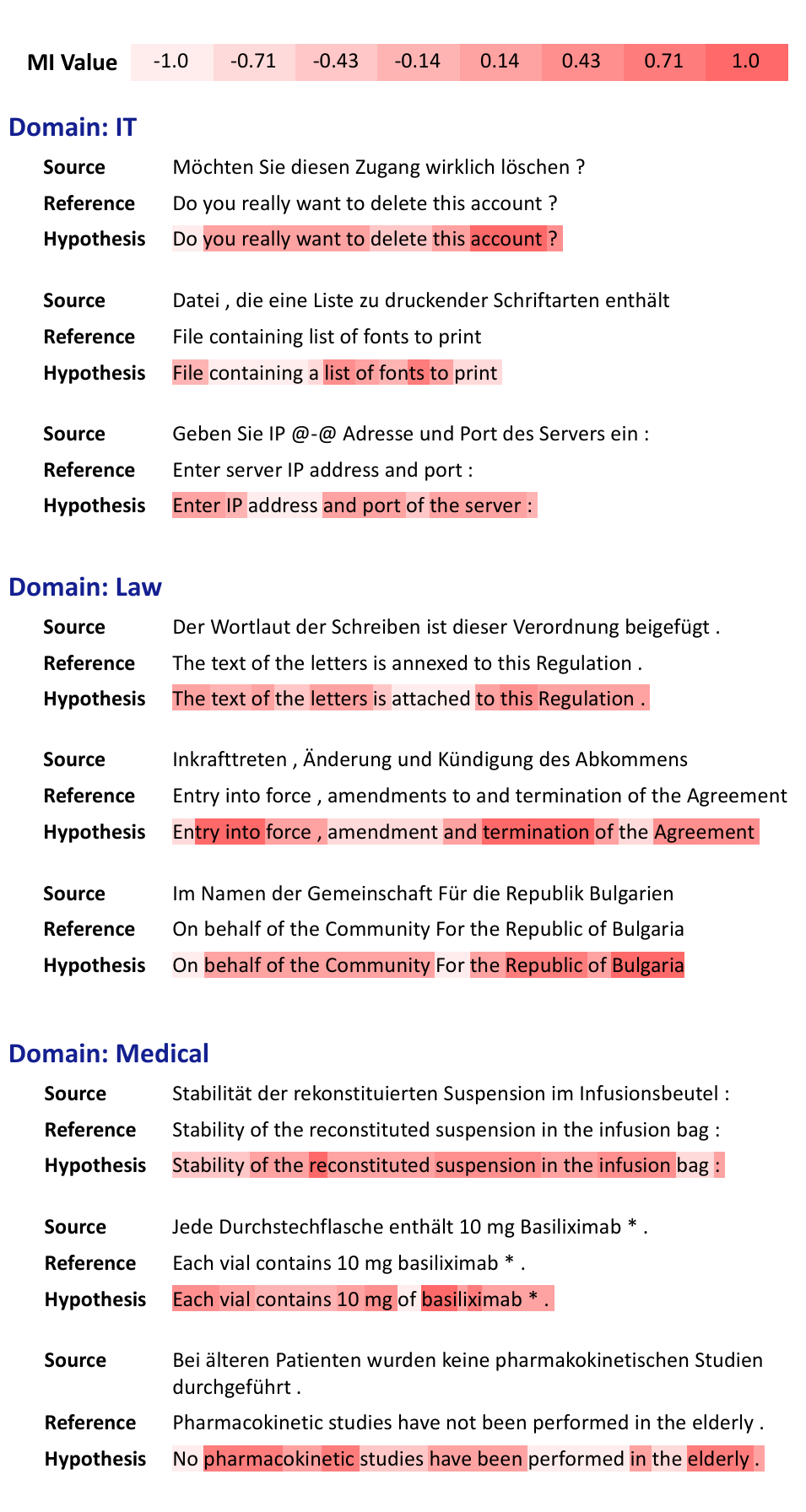}
    \caption{Visualization of examples with MI values from IT, Law, and Medical. Color intensity is correlated with MI value; the more intense the red, the more higher MI value.}
    \label{fig:supp_example_visualization}
\end{figure*}

Figure~\ref{fig:supp_example_visualization} provides more visualizations of test examples with MI values from IT, Law, and Medical.
Color intensity is correlated with MI value; the more intense the red, the more higher MI value.
From the result, domain-specific words (\textit{e.g.}, `account' in IT, `Regulation' in Law, `pharmacokinetic' in Medical) are translated with high MI values.


\end{document}